\documentclass[sn-mathphys,Numbered]{sn-jnl}

\newcommand{\ordiv}[1]{#1}

\usepackage[utf8]{inputenc}%
\usepackage{hyperref}%
\usepackage{float}%
\usepackage{graphicx}%
\usepackage{multirow}%
\usepackage{amsmath,amssymb,amsfonts}%
\usepackage{amsthm}%
\usepackage{mathrsfs}%
\usepackage[title]{appendix}%
\usepackage{xcolor}%
\usepackage{textcomp}%
\usepackage{manyfoot}%
\usepackage{booktabs}%
\usepackage{algorithm}%
\usepackage{algorithmicx}%
\usepackage{algpseudocode}%
\usepackage{listings}%
\usepackage[format=hang, justification=centering]{caption}
\DeclareUnicodeCharacter{2212}{-}

\begin{document}

\title[Article Title]{Synthesis of Batik Motifs using a Diffusion - Generative Adversarial Network}


\author[1]{\fnm{One Octadion} \sur{}}\email{dyonoct@gmail.com}

\author*[1]{\fnm{Novanto Yudistira} \sur{}}\email{yudistira@ub.ac.id}
\equalcont{These authors contributed equally to this work.}

\author[1]{\fnm{Diva Kurnianingtyas} \sur{}}\email{divaku@ub.ac.id}
\equalcont{These authors contributed equally to this work.}

\affil[1]{\orgdiv{Informatics Engineering}, \ordiv{Faculty of Computer Science}, \orgname{Universitas Brawijaya}, \country{Indonesia}}


\abstract{Batik, a unique blend of art and craftsmanship, is a distinct artistic and technological creation for Indonesian society. Research on batik motifs is primarily focused on classification. However, further studies may extend to the synthesis of batik patterns. Generative Adversarial Networks (GANs) have been an important deep learning model for generating synthetic data, but often face challenges in the stability and consistency of results. This research focuses on the use of StyleGAN2-Ada and Diffusion techniques to produce realistic and high-quality synthetic batik patterns. StyleGAN2-Ada is a variation of the GAN model that separates the style and content aspects in an image, whereas diffusion techniques introduce random noise into the data. In the context of batik, StyleGAN2-Ada and Diffusion are used to produce realistic synthetic batik patterns. This study also made adjustments to the model architecture and used a well-curated batik dataset. The main goal is to assist batik designers or craftsmen in producing unique and quality batik motifs with efficient production time and costs. Based on qualitative and quantitative evaluations, the results show that the model tested is capable of producing authentic and quality batik patterns, with finer details and rich artistic variations. The use of the Wasserstein loss function tends to produce batik motifs that are relatively new but less neat than the use of the StyleGAN2-Ada loss. The quality of the dataset also has a positive impact on the quality of the resulting batik patterns. Overall, this research contributes to the integration of Diffusion-GAN technology with traditional arts and culture, especially in the synthesis of batik motifs. However, there is still room for further development in increasing skill and accuracy in producing more detailed batik motifs. The dataset and code can be accessed here: \href{https://github.com/octadion/diffusion-stylegan2-ada-pytorch}{https://github.com/octadion/diffusion-stylegan2-ada-pytorch}}




\maketitle
\section{Introduction}\label{sec1:introduction}

Batik, a combination of art and craftsmanship, is an artistic and technological creation unique to the Indonesian people. It has reached unparalleled levels of design, motifs, and production processes. The deeply meaningful and philosophical batik patterns continue to be explored, drawing inspiration from various customs and cultures in Indonesia. According to the Indonesian Dictionary, a motif refers to a pattern or design that forms diverse and captivating forms \cite{maria2018}.

Research on batik motifs has primarily focused on their classification, aiming to identify specific motifs present in batik fabrics. However, further research can expand into the synthesis of batik motifs. Generating batik patterns automatically can be achieved with the help of artificial intelligence.  intelligence is chosen for its flexibility and applicability in various fields, such as speech recognition, computer vision, natural language processing, and more.

Generative Adversarial Network (GAN) has become a crucial deep learning model in generating synthetic data, such as images or text. Comprising a generator and discriminator, the GAN model collaboratively produces high-quality synthetic data \cite{goodfellow2014generative}.

One challenge in using GAN models lies in their unstable and inconsistent performance during training. Convergence failure or the inability to achieve the desired accuracy level leads to inconsistent outputs. Even when provided with the same input, the model generates outputs of varying quality. This inconsistency stems from the model's struggle to learn consistent patterns in the data or when excessive noise interferes during training, particularly evident when generating complex data like images or videos. Researchers have sought to develop more stable and high-quality models, such as Wasserstein GAN, StyleGAN2, or BigGAN. In this particular case, StyleGAN was chosen due to its implementation simplicity, availability of pre-trained models, and ease of fine-tuning.

StyleGAN2, a variant of the GAN model, excels at generating realistic and high-quality images. The StyleGAN2 architecture adopts a style-based approach, allowing the generator model to separate the style and content aspects of an image. This separation enhances control over the generated image's features at different levels of detail. Nevertheless, further improvements can be made to the StyleGAN model, including the incorporation of the Diffusion technique \cite{karras2019style}.

Inspired by thermodynamic processes, the Diffusion model employs a step-by-step diffusion chain, progressively introducing random noise to the data. The model then learns to reverse this diffusion process, resulting in desired data samples generated from the initial noise \cite{ho2020denoising}.

In the realm of batik application, StyleGAN2 can effectively generate realistic synthetic batik patterns. To achieve optimal performance and enhance the model further, StyleGAN2 can be fine-tuned and adjusted with techniques like the aforementioned diffusion technique. The fine-tuning process for batik involves using carefully curated datasets, encompassing collection, selection, quality determination, and motif choice. Additionally, adjusting the model's architecture, such as modifying the number of layers or kernel size, can improve its ability to learn complex batik patterns. Appropriate loss functions and the incorporation of diffusion techniques should also be considered.

Therefore, the utilization of StyleGAN2, diffusion techniques, and fine-tuning holds promise for generating intricate and realistic synthetic batik patterns of high quality. These patterns can find applications in various fields, including fashion design and other creative industries. They will assist batik designers and artisans in producing unique and top-tier batik patterns, ultimately expediting production time and reducing costs.

\section{Related Works}\label{sec2:related_works}

Goodfellow et al 2014 \cite{goodfellow2014generative} in their research titled "Generative Adversarial Nets", discussed a model with adversarial networks which has two main components: a generator and a discriminator. The generator tries to create data that resemble the original data distribution, while the discriminator attempts to differentiate between the original data and the data produced by the generator. The goal is to train the generator to produce data so convincing that the discriminator cannot distinguish it from the original data. 

Martin Arjovsky, Soumith Chintala, and Leon Bottou  2017 \cite{arjovsky2017wgan}, with their research "Wassertein GAN", they discuss the Wasserstein GAN (WGAN) as an advancement over conventional GANs. WGAN employs the Wasserstein metric in their work to calculate the difference between the generator's distribution and the actual data distribution. Compared to typical GANs, WGAN has shown to be more stable throughout training.

With their research, Ishaan Gulrajani et al. 2017 \cite{gulrajani2017} address difficulties in training GANs and present Gradient Penalty (GP) as a novel solution. It replaces the Weight Clipping approach and demonstrates that GP is more efficient at producing optimal discriminators and delivering superior outcomes across a range of data generation activities. In order to show how GP enhances performance, the paper also examines other training-related topics such overfitting and sample quality measurement.

A study by Fedus et al. (2018) \cite{fedus2018} This research focuses on the use of Generative Adversarial Networks (GANs), experiments are carried out using variations of GANs, such as non-saturating GAN and WGAN-GP, then synthesis trials are carried out to see the effectiveness of each model, using the CelebA, CIFAR-10, and Color MNIST datasets. Based on the results that have been done, it is found that WGAN-GP has the best performance which can overcome the problem of mode collapse.

Tero Karras, Samuli Laine, and Timo Aila 2019 \cite{karras2019style} with his research entitled, "A Style-based Generative Adversarial Network", in which this research focuses on developing GAN models with a style control approach. Their work demonstrates a more stable learning process, eliminates mode collapse, and generates high-quality images. The findings of their research will serve as a foundational methodology for designing GANs in upcoming research. 

Karras et al 2020 \cite{karras2020training} proposed an adaptive discriminator augmentation mechanism for training GANs with limited data. Their experiments demonstrated its effectiveness, even with a small number of training images. The paper discussed implications for image and video authenticity and training efficiency. This study made significant contributions to GAN development with limited data.

Ho et al 2020 \cite{ho2020denoising} introduced Denoising Diffusion Probabilistic Models (DDPMs) as a new generative model that trains a model to predict the original data distribution from a reverse diffusion process. DDPMs have demonstrated impressive sample quality and have inspired various subsequent research in diffusion models.

The research on batik especially Indonesian batik motifs is limited to the classification such as \cite{meranggi2022batik}. Agus Eko Minarno, Moch. Chamdani Mustaqim, et al 2021 \cite{agus2021dcgan}, in their research titled "Deep Convolutional Generative Adversarial Network Application in Batik Pattern Generator", discuss the application of the Deep Convolutional method and GANs in generating batik patterns. Their research findings demonstrate that batik patterns can be effectively produced using the GAN method.  In this work we try to explore the possibility of generating various Indonesian batik motifs using diffusion model and GANs.

\section{Research Method and Materials}\label{sec3:method}

\subsection{Generative Adversarial Network}\label{subsec2:gan}
The Generative Adversarial Network (GAN) is a neural network architecture used for generating synthetic data that appears to be taken from the original distribution. A GAN consists of two opposing models, which is a generator ($G$) and a discriminator ($D$) \cite{goodfellow2014generative}.

The generator is a neural network that generates synthetic data. The generator receives random noise $z$ as input and converts it into an expected output as synthetic data $G(z)$.

The discriminator is a neural network used to distinguish between original data and synthetic data generated by the generator. The discriminator receives input data (either original or synthetic) and outputs a score $D(x)$ or $D(G(z))$ indicating how convincing the data is considered to be original.

The training of the GAN involves optimizing the generator to fool the discriminator into not being able to distinguish synthetic data from original data, and optimizing the discriminator to accurately distinguish synthetic data from original data. This process is repeated until convergence.

This can be represented by the following loss function:
\begin{equation}
\min_{G} \max_{D} V(D,G) = \mathbb{E}_{x\sim p_{\text{data}}(x)}[\log D(x)] + \mathbb{E}_{z\sim p_{z}(z)}[\log(1-D(G(z)))]
\end{equation}
Where $\mathbb{E}$ is the symbol for expectation or average, $x$ is the data from the original distribution $p_{\text{data}}$, $x$ is the data from the original distribution $p_{\text{data}}$, $D(x)$ is the discriminator's belief that $x$ is real, $G(z)$ is the synthetic data generated by the generator, and $G(z)$ is the synthetic data generated by the generator.
The Discriminator ($D$) is trained to maximize this loss function (trying to make $D(x)$ close to 1 and $D(G(z))$ close to 0), while the Generator ($G$) is trained to minimize the second part of this loss function (trying to make $D(G(z))$ close to 1).
\subsection{Style-GAN}\label{subsec2:stylegan}
The Style Generative Adversarial Network (StyleGAN) is an advancement from previous GANs. Unlike its predecessors, StyleGAN can generate new images with significantly higher quality \cite{karras2019style}.

StyleGAN, like previous GANs, employs the concept of adversarial training, where the training consists of two competing networks: the generator and the discriminator. The generator is responsible for creating fake images, while the discriminator is tasked with distinguishing between real and fake images. In this process, both networks train each other, allowing the generator to create images that increasingly fool the discriminator until the generator produces images that resemble the real ones.

There are several components that distinguish StyleGAN from traditional GANs, which make the StyleGAN model superior to traditional GANs. They include: Mapping Network - a mapping concept to convert latent vectors into style vectors that govern various visual aspects, Weight Demodulation - a technique used to remove distortions in the generated images, Noise Injection - a noise injection into the generator to influence details and textures so that images are more varied.

\subsection{Diffusion Model}\label{subsec2:diffusion}

Diffusion models are a type of generative model used to generate new data similar to the data used in training. Fundamentally, diffusion models work by degrading the training data through repeated addition of Gaussian noise in its iterations, and then the model learns how to restore the noisy data by reversing the noise addition process. After training the model, new data can be generated just by feeding random noise into the learned noise-removal process \cite{ho2020denoising}.

\begin{figure}[H]
\centering
\includegraphics[width=0.6\textwidth]{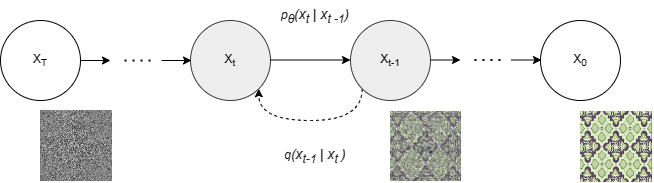}
\caption{Reverse process in diffusion model.}
\label{fig:reverse_diffusion_model}
\end{figure}

More specifically, the diffusion model is a latent variable model that maps to the latent space using a Markov chain. This chain progressively adds noise to the data to obtain a posterior close to $q(x_{1:T}|x_0)$, where $x_{1:T}$ are latent variables of the same dimension as $x_0$. As illustrated in the Figure \ref{fig:adding_diffusion_model} below, in a process from the Markov chain, an image will be added with Gaussian noise until it becomes pure noise, and then the diffusion model's task is to learn to reverse the process as shown in the Figure \ref{fig:reverse_diffusion_model}.
\begin{figure}[ht]
\centering
\includegraphics[width=0.6\textwidth]{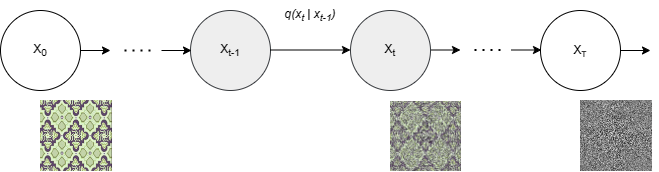}
\caption{Adding noise process in diffusion model.}
\label{fig:adding_diffusion_model}
\end{figure}
\subsection{Proposed Method}\label{subsec2:proposed}
\subsubsection{Diffusion-GAN}\label{subsec3:diffgan}
Diffusion-GAN is an innovative framework for Generative Adversarial Networks (GANs) that introduces a unique approach using a forward diffusion chain to generate instant noise distributed by a Gaussian mixture. Diffusion-GAN comprises three components: an adaptive diffusion process, a diffusion time step-dependent discriminator, and a generator \cite{zhendong2022}.
\begin{enumerate}
  \item The same adaptive diffusion mechanism is used to diffuse both the observed data and the created data. Each diffusion time step uses a different noise-to-data ratio. The mathematical definition of this diffusion process is as follows:
\begin{equation}
q(x_1:T | x_0) := \prod_{t=1}^{T} q(x_t | x_{t−1}),
\end{equation}

\begin{equation}
q(x_t | x_{t−1}) := N(x_t; \sqrt{1 − \beta_t} x_{t−1}, \beta_t \sigma^2 I).
\end{equation}
  \item Diffusion Time Step Discriminator: Between diffused real data and diffused produced data, the discriminator is trained to distinguish. The diffusion time step has an impact on how well the discriminator performs.
  \item  Generator: By backpropagating through the forward diffusion chain, the generator picks up information from the feedback of the discriminator. To balance the amounts of noise and data, the diffusion chain's length is dynamically changed.
\end{enumerate}
When utilizing Diffusion-GAN, an actual image or a generated image can be used as the diffusion process input. The several stages of the diffusion process gradually magnify the noise in the image. The generator and the data affect how many diffusion steps there are. The diffusion process must be differentiable in order to calculate the derivative of the output from the input. As a result, it is possible to update the generator by diffusing the gradient from the discriminator.
Diffusion-GAN compares noisy versions of actual and created images that were obtained by sampling from a Gaussian mixture distribution over the diffusion steps using the time step-dependent discriminator, as opposed to typical GANs that directly compare real and generated images. Due to different noise-to-data ratios, this distribution demonstrates the property that certain components contribute more noise than others. Sampling from this distribution has two benefits: first, by addressing the problem of vanishing gradients, it stabilizes the training process; second, it enriches the dataset by producing multiple noisy copies of the same image, improving data efficiency and generator diversity.
\subsubsection{Architecture}\label{subsec3:architecture}
\begin{figure}[H]
\centering
\includegraphics[width=0.7\textwidth]{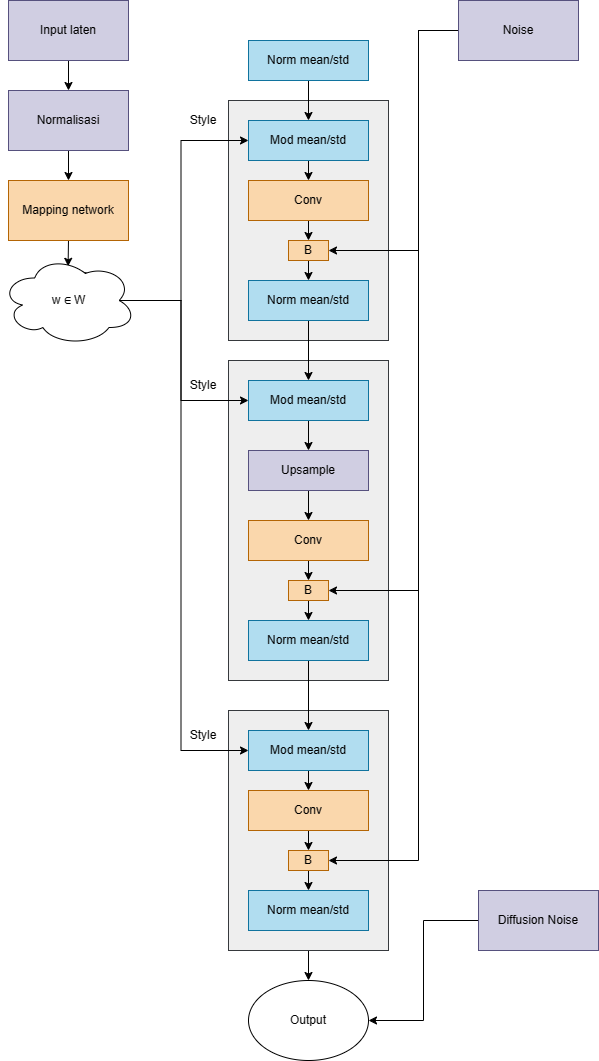}
\caption{An overview of architecture.}
\label{fig:approach_architecture}
\end{figure}
As per the Figure \ref{fig:approach_architecture} above, we utilize the StyleGAN2-Ada architecture as the primary architecture, which comprises a mapping network made up of several fully connected layers. Following this, the generator consists of several components, namely a generation block for each resolution level, a style module, the progressive growing technique which starts training at a low resolution and gradually increases the resolution during the training process, and the implementation of weight demodulation to reduce emerging artifacts. Subsequently, the discriminator is composed of several components, including a discrimination block for each resolution level, downsampling, and a fully-connected layer to produce a final score indicating whether the input image is considered real or synthetic. Lastly, we modify the architecture by adding a diffusion function to inject Gaussian noise into the images generated during training, enabling the discriminator to learn to distinguish noise in real images and noise in fake images, thereby enhancing the model, where details are shown in the Figure \ref{fig:detail_diffusion} below.
\begin{figure}[H]
\centering
\includegraphics[width=0.8\textwidth]{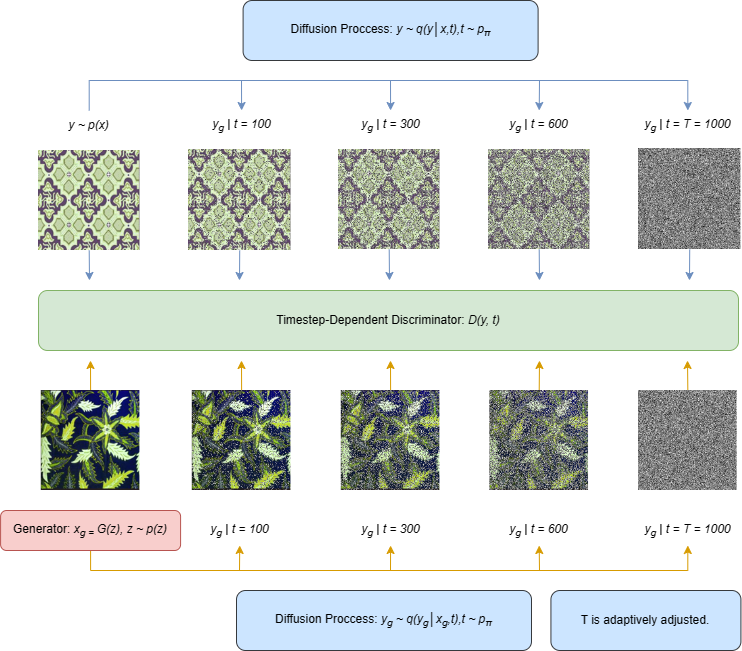}
\caption{Detail of Diffusion Proccess.}
\label{fig:detail_diffusion}
\end{figure}

\subsubsection{Algorithm}\label{subsec3:algorithm}
Below is the algorithm of Diffusion-GAN:
\begin{algorithm}[H]
\caption{Diffusion-GAN}
\label{alg:diffusion-gan}
\begin{algorithmic}[1]
\State $i \gets 0$
\While{$i \leq$ number of training iterations}
    \State \textbf{Step I: Update discriminator}
    \State Sample minibatch of $m$ noise samples $\{z_1, z_2, \ldots, z_m\}$ from the distribution $p_z(z)$.
    \State Generate samples $\{x_{g,1}, x_{g,2}, \ldots, x_{g,m}\}$ using the generator $G$ with inputs $\{z_1, z_2, \ldots, z_m\}$.
    \State Sample minibatch of $m$ data examples $\{x_1, x_2, \ldots, x_m\}$ from the distribution $p(x)$.
    \State Sample $\{t_1, t_2, \ldots, t_m\}$ uniformly with replacement from a given tepl list.
    \For{$j \in \{1, 2, \ldots, m\}$}
        \State Sample $y_j \sim q(y_j |x_j , t_j )$ and $y_{g,j}$ from the distribution $q(y_{g,j} |x_{g,j} , t_j )$.
    \EndFor
    \State Update discriminator by maximizing $L_D = -E_{x \sim p_{data}(x)}[\log D(x)] - E_{z \sim p_{z}(z)}[\log(1 - D(G(z)))]$
    
    \State \textbf{Step II: Update generator}
    \State Sample minibatch of $m$ noise samples $\{z_1, z_2, \ldots, z_m\}$ from the distribution $p_z(z)$.
    \State Generate samples $\{x_{g,1}, x_{g,2}, \ldots, x_{g,m}\}$ using the generator $G$ with inputs $\{z_1, z_2, \ldots, z_m\}$.
    \State Sample $\{t_1, t_2, \ldots, t_m\}$ with replacement from a given tepl list.
    \For{$j \in \{1, 2, \ldots, m\}$}
        \State Sample $y_{g,j}$ from the distribution $q(y_{g,j} |x_{g,j} , t_j )$.
    \EndFor
    \State Update generator by minimizing $L_G = E_{z \sim p_{z}(z)}[\log(1 - D(G(z)))]$
    
    \State \textbf{Step III: Update diffusion}
    \If{$i \mod 4 == 0$}
        \State Calculate $r_d = E_{y,t \sim p(y,t)} [\text{sign}(D_{\phi}(y, t) - 0.5)]$ and update $T = T + \text{sign}(r_d - d_{\text{target}}) \cdot C$.
        \State Sample the tepl list with tepl $= [0, \ldots , 0, t_1, \ldots , t_{32}]$, where $t_k \sim p_{\pi}$ for $k \in \{1, \ldots , 32\}$, and $t \sim p_{\pi}:= (\text{uniform: Discrete, priority: Discrete})$.
    \EndIf
    \State $i \gets i + 1$
\EndWhile
\end{algorithmic}
\end{algorithm}

The Algorithm \ref{alg:diffusion-gan} above is the algorithm used in training the Generative Adversarial Network model with the Diffusion method. In each training iteration, the following steps are performed: First, the discriminator is updated to maximize the Discriminator's loss function. First, we take $m$ minibatch samples of noise samples $z_1, z_2, \ldots, z_m$ from the $p_z(z)$ distribution. Then, generator $G$ is used to generate samples $x_{g,1}, x_{g,2}, \ldots, x_{g,m}$ using inputs $z_1, z_2, \ldots, z_m$. Next, we take $m$ minibatch samples from the sample data $x_1, x_2, \ldots, x_m$ from the $p(x)$ distribution. The samples $t_1, t_2, \ldots, t_m$ are taken randomly by replacement from the list of tepls given. Next, for each $j$ in $1, 2, \ldots, m$, sample $y_j$ from the distribution $q(y_j |x_j , t_j )$ and $y_{g,j}$ from the distribution $q(y_{g,j} |x_{g,j} , t_j )$. The discriminator is updated by maximizing the log likelihood of the samples using the Discriminator loss function.

Next, the Generator is updated by minimizing the Generator loss function. First, we take $m$ minibatch samples of noise samples $z_1, z_2, \ldots, z_m$ from the $pz(z)$ distribution. Generator $G$ is used to generate samples $x_{g,1}, x_{g,2}, \ldots, x_{g,m}$ using the input $z_1, z_2, \ldots, z_m$. Samples $t_1, t_2, \ldots, t_m$ are taken with replacement from the list of tepls given. For each $j$ in $1, 2, \ldots, m$, sample $y_{g,j}$ from the distribution $q(y_{g,j} |x_{g,j} , t_j )$. The generator is updated by minimizing the log likelihood of the samples using the Generator loss function.

And last, if $i$ modulus 4 is equal to 0, then the following steps are taken. First, $r_d$ is calculated by taking the average sign of the difference between $D_{\phi}(y, t)$ and 0.5, where $D$ is the discriminator and $\phi$ is the internal representation of the data. The $r_d$ value is used to update the $T$ value by taking the sign of the difference between $r_d$ and $dtarget$, then multiplied by $C$ and added to $T$. Next, the tepl list is updated by selecting $t$ randomly from the $p_{\pi}$ distribution. The $p_{\pi}$ distribution consists of a uniform discrete distribution for elements 0 and a priority discrete distribution for $t$ elements in tepl. After the steps are completed, the initialization value is incremented by 1 and returns to step 2 to continue iteration and training repetition. This algorithm describes how discriminators and generators interact with each other in Diffusion-GAN training to produce a model capable of producing new data samples that resemble the original data.

\subsection{Data Acquisition and Preprocessing}\label{subsec2:data}
\begin{figure}[H]
\centering
\begin{minipage}[b]{0.3\textwidth}
\centering
\includegraphics[width=\textwidth]{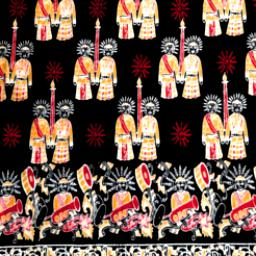}
\caption*{(a)} 
\end{minipage}
\hfill
\begin{minipage}[b]{0.3\textwidth}
\centering
\includegraphics[width=\textwidth]{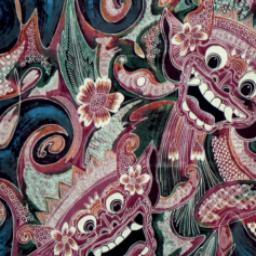}
\caption*{(b)} 
\end{minipage}
\hfill
\begin{minipage}[b]{0.3\textwidth}
\centering
\includegraphics[width=\textwidth]{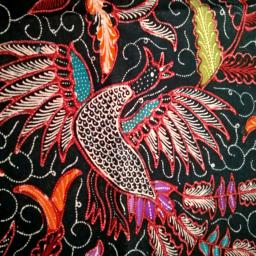}
\caption*{(c)} 
\end{minipage} 
\vspace{10pt}
\caption{Preprocessed dataset samples : (a) Batik Betawi, (b) Batik Bali, (c) Batik Cendrawasih}
\label{fig:preprocessed_sample}
\end{figure}

The batik motif image data used in this research was sourced from various platforms such as the internet, Kaggle, and GitHub \cite{gultom2018batik}. There are 20 types of batik motifs used in the study, namely: nitik, bali, ceplok, lereng, kawung, megamendung, parang, pekalongan, priangan, sidomukti, betawi, cendrawasih, ciamis, garutan, gentongan, keraton, lasem, sekar, sidoluhur, and tambal, we chose 20 types of batik with several considerations such as data availability and dataset diversity, which is because the models require diverse data to be able to learn well. Figure \ref{fig:comparison_data} below shows a comparison of the data before and after preprocessing, the first image shows the size of the image which is still not the same as the other images, besides that the orientation of the image and color are still original, where in the second image the size of the image is in accordance with changes in orientation and color alteration, Next, the Figure \ref{fig:preprocessed_sample} shows sample images that have been pre-processed, there are examples of Betawi, Balinese and Cendrawasih batik.

\begin{figure}[H]
\centering
\begin{minipage}[b]{0.3\textwidth}
\centering
\includegraphics[width=\textwidth]{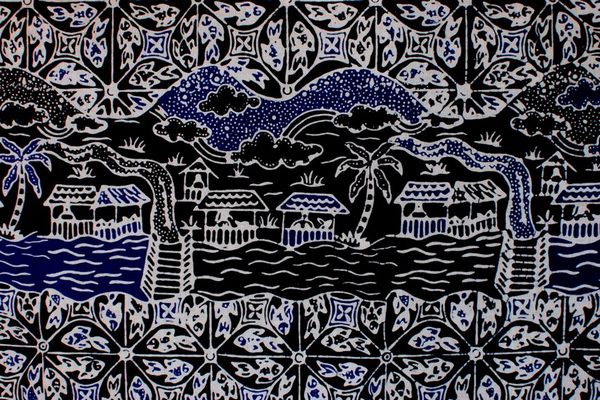}
\caption*{(a)} 
\end{minipage}
\hfill
\begin{minipage}[b]{0.3\textwidth}
\centering
\includegraphics[width=\textwidth]{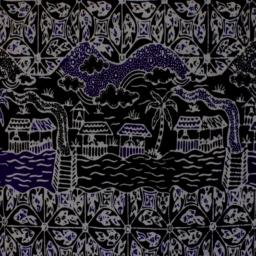}
\caption*{(b)} 
\end{minipage}
\vspace{10pt}
\caption{Comparison between unpreprocessed and preprocessed data sample: (a) Unpreprocessed Data, (b)
Preprocessed Data}
\label{fig:comparison_data}
\end{figure}
For each type of batik, images were selected with clearly visible motifs, while images with excessive noise, non-batik shapes, and unclear motifs were removed from the dataset. Subsequently, data augmentation was performed on the dataset, as GAN models require as much data as possible to achieve good performance. Given that each type of batik has a different number of data, the amount of data augmentation was adjusted based on the number of data for each type of batik. Augmentation was done by cropping, flipping horizontally and vertically, and color alteration. We aimed to balance the number of data for each type of batik to avoid overfitting, which could result in one type of batik dominating the generated results. Therefore, augmentation for each type of batik was performed with 1000 data, with the total data reaching 20.000 batik motif data. But, at first, we try to train with 10.000 data, to see how the data affects the results.
\section{Result and Discussion}\label{sec4:result}
\subsection{Hyperparameters}\label{subsec2:parameter}
In the baseline StyleGAN model that will be used, there are several hyperparameters that can be set to achieve optimal results. In this experiment, with 256x256 resolution batik image data, the training configuration consists of the following parameters:

\begin{itemize}
    \item \texttt{ref\_gpus}: The number of reference GPUs used during training. The number of reference GPUs can affect the mini-batch size used during training.
    \item \texttt{kimg}: The total number of training iterations in thousands.
    \item \texttt{mb} (mini-batch): The number of samples processed in a single training iteration.
    \item \texttt{mbstd} (mini-batch standard deviation): A factor used to control the variation in the mini-batch samples.
    \item \texttt{fmaps}: The factor determining the number of feature maps used in the model. This value can affect the complexity and capacity of the model.
    \item \texttt{lrate} (learning rate): The rate at which the model's weights are updated during training.
    \item \texttt{gamma}: A factor controlling stability during training.
    \item \texttt{ema} (Exponential Moving Average): A technique used to smooth the model during training and achieve better results.
    \item \texttt{ramp}: The gradual increase value used during training.
    \item \texttt{map}: The factor determining the number of attribute maps used in the model.
\end{itemize}

Following the \texttt{paper256} configuration \cite{karras2019style}, we have set the values as follows: \texttt{ref\_gpus=8}, \texttt{kimg=25000}, \texttt{mb=64}, \texttt{mbstd=8}, \texttt{fmaps=0.5}, \texttt{lrate=0.0025}, \texttt{gamma=1}, \texttt{ema=20}, \texttt{ramp=None}, and \texttt{map=8}.

After hyperparameter in StyleGAN, in Diffusion there are several hyperparameters that introduced, the following parameters are \texttt{dtarget} which is a threshold to identify whether the current discriminator is overfitting, \texttt{$p_{\pi}$} which is the sampling distribution, and \texttt{noise}. in this experiment we set the \texttt{dtarget=0.6}, \texttt{$p_{\pi}$=priority} and \texttt{noise=0.05}.

\subsection{Architecture Comparison}\label{subsec2:comparison}
First, we experiment with parameters as mentioned in the previous chapter and with 10,000 datasets, Table \ref{tab:table1} shows the best results are produced by the baseline+diffusion, with the best FID value around 45.611, and KID around 0.0084. The best precision and recall results are obtained from the baseline+diffusion and the baseline that uses Wasserstein loss with a precision value of 0.269 and a recall of 0.1414.
\begin{table}[ht]
\caption{Experiment with 10.000 data results, \textbf{Bold} number show
best score while \underline{underline} shows second bests}\label{tab:table1}
\begin{tabular*}{\textwidth}{@{\extracolsep\fill}lcccccc}
\toprule%
Model & FID $\downarrow$ & KID $\downarrow$ & Prec $\uparrow$  & Rec $\uparrow$ \\
\midrule
base  & 65.699 & 0.0153  &  0.249  &  0.0573\\
base$+$Wloss  & 57.123 & 0.0180  &  0.251   & \textbf{0.1414} \\
base$+$Diffusion  &\textbf{45.611} & \textbf{0.0084}  &  \textbf{0.269} &  0.0785\\
base$+$Wloss$+$Diffusion  & \underline{48.297} & \underline{0.0135}  &  \underline{0.254} & \underline{0.1391}\\
\botrule
\end{tabular*}
\end{table}

Next, we tried to retrain with the addition of the dataset to 20,000 data, to see if there is an increase in performance and better results. Table \ref{tab:table2} shows the best results are obtained by the baseline+diffusion with the best FID value around 29.045 and KID around 0.00643, the best precision and recall results are obtained by the baseline+diffusion with Wasserstein loss with a precision value of 0.1744 and a recall from base+wassertein loss with 0.1612. From the experiments that have been conducted, it is known that the addition of the dataset is able to improve performance with better results.
\begin{table}[ht]
\caption{Experiment with 20.000 data results, \textbf{Bold} number show
best score while \underline{underline} shows second bests}\label{tab:table2}
\begin{tabular*}{\textwidth}{@{\extracolsep\fill}lcccccc}
\toprule%
Model & FID $\downarrow$ & KID $\downarrow$ & Prec $\uparrow$  & Rec $\uparrow$ \\
\midrule
base  & 46.799 & 0.01287  &  0.1590  &  0.1117\\
base$+$Wloss  & 48.781 & 0.01800  &  0.1566   & \textbf{0.1612}\\
base$+$Diffusion  &\textbf{ 29.045} & \textbf{0.00643}  &  \underline{0.1628} &  0.125\\
base$+$Wloss$+$Diffusion  & \underline{36.756} & \underline{0.01073}  &  \textbf{0.1744} &\underline{0.1426}\\
\botrule
\end{tabular*}
\end{table}

Lastly, we attempted to retrain the model using a combination of the previous 20,000 batik motif dataset and approximately 1,200 batik motifs from the ITB-mBatik dataset \cite{chrys2023}. The ITB-mBatik dataset consists of unique, high-quality, digitally created symmetric batik patterns, which differ from the original batik motif images in the previous dataset. We trained the model with this combined dataset to see if it could generate even more unique and diverse motifs after the addition of unique motifs from the new dataset.

As seen in Table \ref{tab:table3}, the best FID score is achieved by the baseline+diffusion model with a value of approximately 33.0104, slightly higher than the previous training. Similarly, the best KID score is also obtained by the baseline+diffusion model with a value of around 0.006705. The precision and recall values are obtained by the baseline+diffusion and baseline+wloss models, with values of approximately 0.1777 and 0.16861, respectively.
\begin{table}[ht]
\caption{Experiment with combined 20.000 data and ITB-mBatik data results, \textbf{Bold} number show
best score while \underline{underline} shows second bests}\label{tab:table3}
\begin{tabular*}{\textwidth}{@{\extracolsep\fill}lcccccc}
\toprule%
Model & FID $\downarrow$ & KID $\downarrow$ & Prec $\uparrow$  & Rec $\uparrow$ \\
\midrule
base  & 57.4302 & 0.01689  &  0.1315  &  0.08601\\
base$+$Wloss  & \underline{42.9192} & \underline{0.01423}  &  0.16889   & \textbf{0.16861}\\
base$+$Diffusion  &\textbf{ 33.0104} & \textbf{0.006705}  &  \textbf{0.1777} &  0.1172\\
base$+$Wloss$+$Diffusion  & 43.4331 & 0.01550  &  \underline{0.1756} &\underline{0.1372}\\
\bottomrule
\end{tabular*}
\end{table}
\subsection{Image Results}\label{subsec2:image}
Here are the results from the tested models. According to our evaluation, the outputs generated using Wasserstein loss exhibited more unique motifs compared to those without it. However, these motifs tended to lack organization and neatness. On the other hand, the model employing the Diffusion-GAN method demonstrated the best performance, both in terms of FID scores and the visual quality of the generated images.
\begin{figure}[H]
\centering
\includegraphics[width=0.9\textwidth]{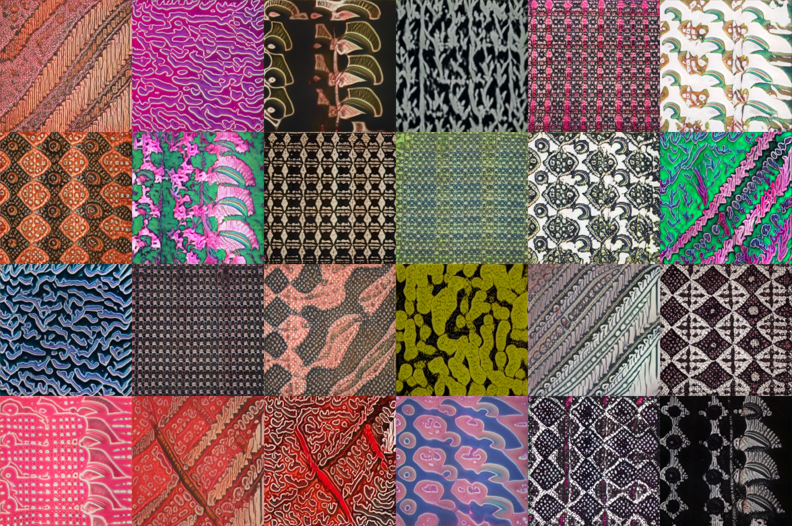}
\caption{Baseline results (FID 46.799).}
\label{fig:results_baseline}
\end{figure}
Using the baseline model, the generated batik images produced by this method as shown in Figure \ref{fig:results_baseline} exhibit vibrant and high-quality colors, as well as neat and well-defined patterns. The batik motifs are clearly visible, and the details are accurately depicted. One of the strengths of this method is its ability to preserve the authenticity of the colors and shapes of the original batik motifs while maintaining the stability of the generated patterns. However, a limitation of this method is its limited capability to generate completely new motifs. Although the generated motifs may possess new and distinct patterns from the original data, they tend to resemble existing batik patterns.
\begin{figure}[H]
\centering
\begin{minipage}[b]{0.4\textwidth}
\centering
\includegraphics[width=\textwidth]{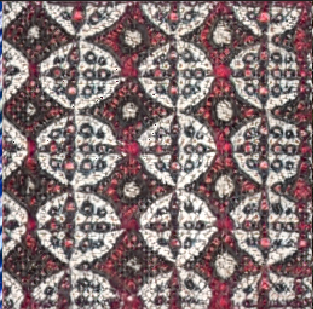}
\caption*{(a)} 
\end{minipage}
\hfill
\begin{minipage}[b]{0.4\textwidth}
\centering
\includegraphics[width=\textwidth]{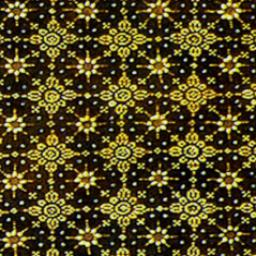}
\caption*{(b)} 
\end{minipage}
\vspace{10pt}
\caption{Comparison between generated sample and real data: (a) Generated Sample, (b)
Real Data}
\label{fig:baseline_comparison}
\end{figure}
As illustrated in the Figure \ref{fig:baseline_comparison} above, the generated sample showcase some additional new patterns, yet they still bear a resemblance to the original batik style, namely the Nitik motifs. However, it is important to note that the motifs in the samples represent fresh and distinct variations compared to the original dataset.
\begin{figure}[H]
\centering
\includegraphics[width=0.9\textwidth]{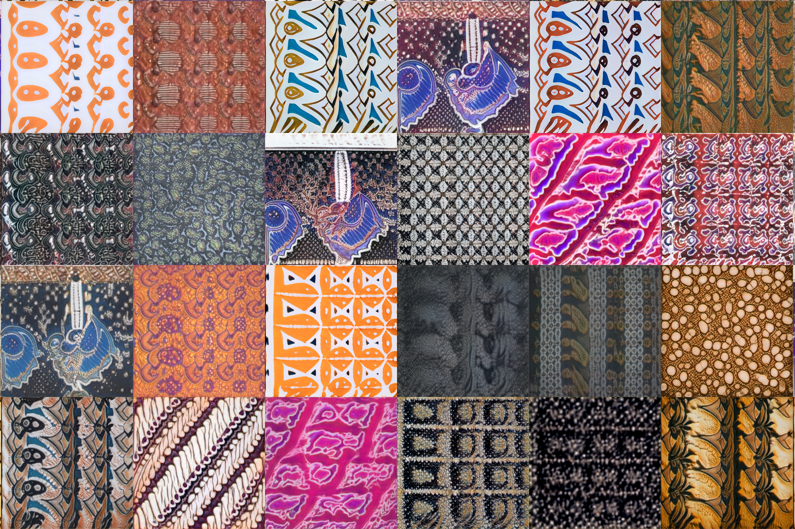}
\caption{Baseline results with combined datasets (FID 57.4302).}
\label{fig:results_baseline_combined}
\end{figure}
The Figure \ref{fig:results_baseline_combined} above represents the outcomes of training the baseline model using a combined dataset, showcasing an enhancement in uniqueness and the emergence of new diverse patterns in the generated samples.
\begin{figure}[H]
\centering
\includegraphics[width=0.9\textwidth]{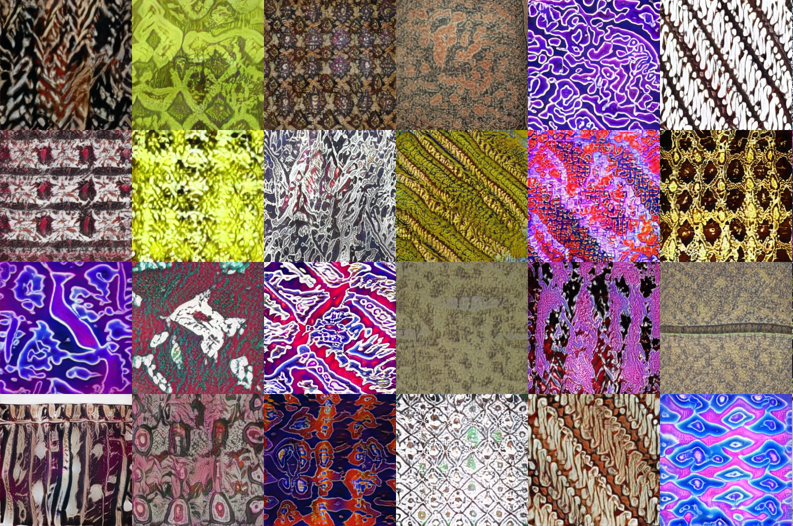}
\caption{Baseline+Wassertein loss results (FID 48.781).}
\label{fig:results_baseline+wloss}
\end{figure}
Using the baseline model with Wasserstein loss, the generated batik images produced by this method as shown in Figure \ref{fig:results_baseline+wloss} exhibit new and distinct patterns compared to the original dataset. Additionally, the arrangement of patterns forming the motifs also differs from the original data. This method excels in generating novel pattern shapes and their arrangement within the motifs. However, a limitation of this method is its reduced ability to generate neatly organized motifs, as the patterns within the motifs may appear somewhat random.
\begin{figure}[H]
\centering
\begin{minipage}[b]{0.4\textwidth}
\centering
\includegraphics[width=\textwidth]{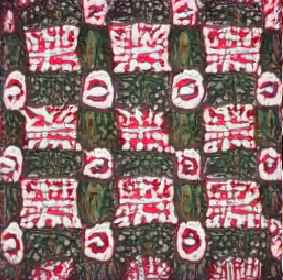}
\caption*{(a)} 
\end{minipage}
\hfill
\begin{minipage}[b]{0.4\textwidth}
\centering
\includegraphics[width=\textwidth]{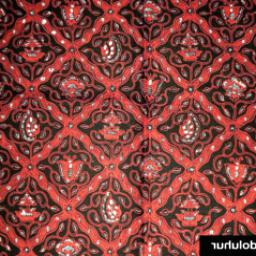}
\caption*{(b)} 
\end{minipage}
\vspace{10pt}
\caption{Comparison between generated sample and real data: (a) Generated Sample, (b)
Real Data}
\label{fig:baseline+wloss_comparison}
\end{figure}
In Figure \ref{fig:baseline+wloss_comparison}, the generated samples exhibit motifs that are relatively new, incorporating elements from batik styles such as Sidoluhur, Nitik, and Kawung. However, it is also apparent that the patterns in these samples appear random and lack organization.
\begin{figure}[H]
\centering
\includegraphics[width=0.9\textwidth]{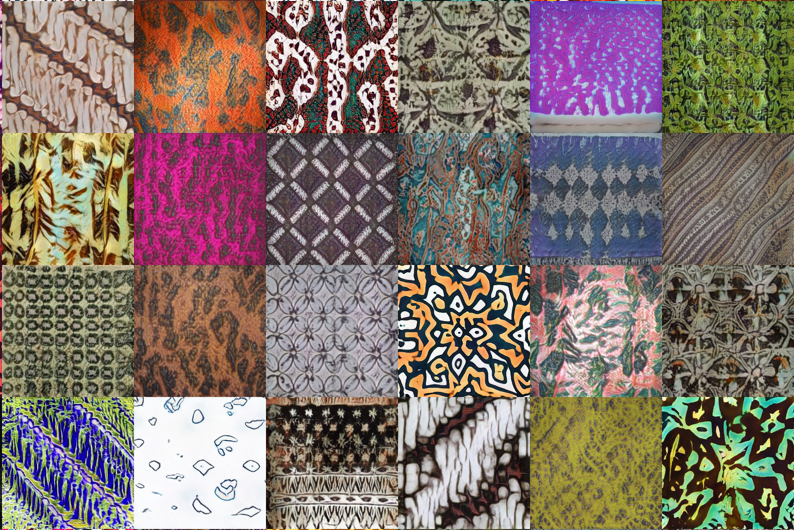}
\caption{Baseline+WLoss results with combined datasets (FID 42.9192).}
\label{fig:results_baseline+wloss_combined}
\end{figure}
The results of the baseline model utilizing Wasserstein loss as shown in Figure \ref{fig:results_baseline+wloss_combined} are depicted, revealing a more organized arrangement of motifs in the generated output compared to the utilization of the previous dataset.
\begin{figure}[H]
\centering
\includegraphics[width=0.9\textwidth]{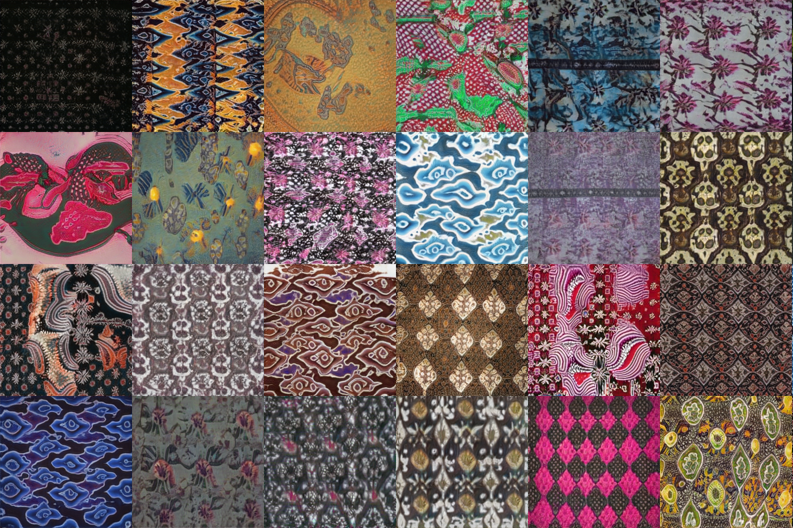}
\caption{Baseline+Diffusion results (FID 29.045).}
\label{fig:results_baseline+diffusion}
\end{figure}
By incorporating the diffusion technique into the baseline model, the resulting batik images as shown in Figure \ref{fig:results_baseline+diffusion} exhibit a wide range of vibrant and diverse colors, accompanied by novel patterns that differ from the original dataset. The arrangement of these patterns forming new motifs is well-structured and orderly. This method excels in generating high-quality, fresh, and neatly organized new motifs. However, a limitation of this approach is the presence of batik motifs that still resemble the original data, as not all generated motifs attain the same level of high-quality novelty.
\begin{figure}[H]
\centering
\begin{minipage}[b]{0.4\textwidth}
\centering
\includegraphics[width=\textwidth]{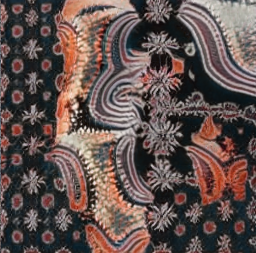}
\caption*{(a)} 
\end{minipage}
\hfill
\begin{minipage}[b]{0.4\textwidth}
\centering
\includegraphics[width=\textwidth]{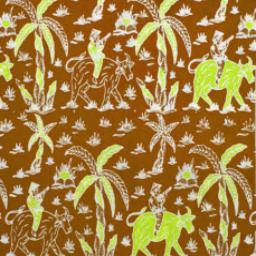}
\caption*{(b)} 
\end{minipage}
\vspace{10pt}
\caption{Comparison between generated sample and real data: (a) Generated Sample, (b)
Real Data}
\label{fig:base+diffusion_comparison}
\end{figure}
In Figure \ref{fig:base+diffusion_comparison} shows the results portray fresh and new motifs, subtly combining elements from Betawi and Cendrawasih batik patterns, resulting in an elegantly arranged motif within the sample.
\begin{figure}[H]
\centering
\includegraphics[width=0.9\textwidth]{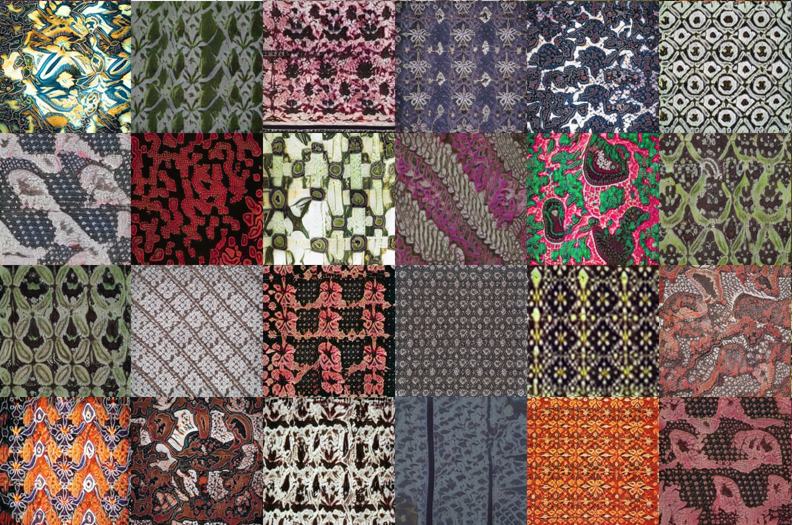}
\caption{Baseline+Diffusion results with combined datasets (FID 33.0104).}
\label{fig:results_baseline+diffusion_combined}
\end{figure}
Figure \ref{fig:results_baseline+diffusion_combined} showcases the generated samples from the baseline+diffusion model. When employing the diffusion method, it is evident that the results tend to exhibit more intricate patterns compared to the model without diffusion. This is attributed to the injection of noise during training, enabling the model to gain a better understanding of image motifs.
\begin{figure}[H]
\centering
\includegraphics[width=0.9\textwidth]{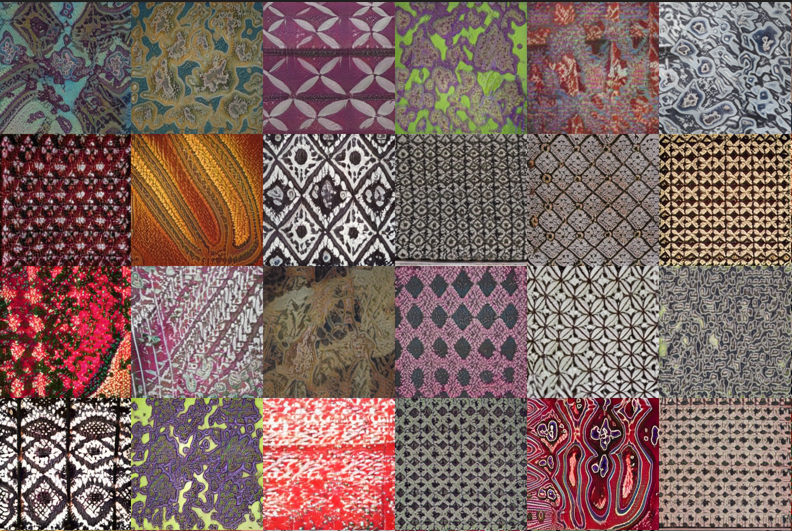}
\caption{Baseline+Diffusion+Wassertein loss results (FID 36.756).}
\label{fig:results_baseline+diffusion+wloss}
\end{figure}
The last approach is the baseline model with the addition of diffusion technique and the utilization of Wasserstein loss. The resulting batik images using this method as shown in Figure \ref{fig:results_baseline+diffusion+wloss} share similarities with the previous diffusion-based approach in terms of vibrant colors and fresh new patterns. However, similar to the previous Wasserstein loss method, the generated patterns tend to be random and lack organization. This method excels in producing high-quality, innovative motifs where the patterns created are fresh and new. A drawback of this approach is the lack of orderly arrangement and the presence of randomness in the generated patterns.
\begin{figure}[H]
\centering
\begin{minipage}[b]{0.4\textwidth}
\centering
\includegraphics[width=\textwidth]{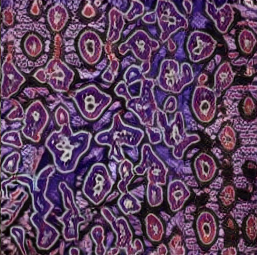}
\caption*{(a)} 
\end{minipage}
\hfill
\begin{minipage}[b]{0.4\textwidth}
\centering
\includegraphics[width=\textwidth]{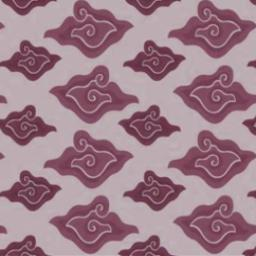}
\caption*{(b)} 
\end{minipage}
\vspace{10pt}
\caption{Comparison between generated sample and real data: (a) Generated Sample, (b)
Real Data}
\label{fig:base+diffusion+wloss_comparison}
\end{figure}
In the last model's results as depicted in the Figure \ref{fig:base+diffusion+wloss_comparison} above, the generated samples showcase fresh and new motifs. However, the resulting patterns also lack organization and appear random. On the other hand, in the results obtained using the combined dataset, as observed in Figure \ref{fig:results_baseline+diffusion+wloss_combined} below, the motifs in the images appear more organized compared to before.
\begin{figure}[H]
\centering
\includegraphics[width=0.9\textwidth]{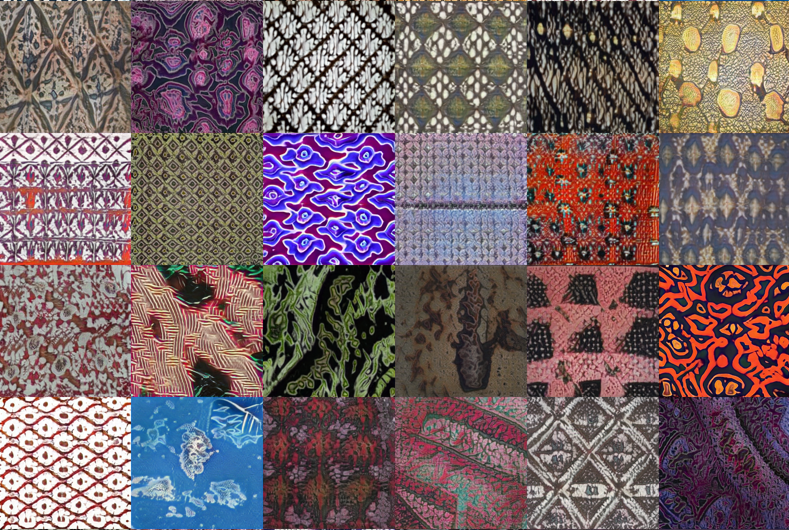}
\caption{Baseline+Diffusion+WLoss results with combined datasets (FID 43.4331).}
\label{fig:results_baseline+diffusion+wloss_combined}
\end{figure}
\subsection{Analysis}\label{subsec2:analysis}
From the tests that have been conducted, evaluation is carried out using several metrics such as FID, KID, Precision and Recall. Images are judged based on the quality and diversity or variation of the resulting new motifs. On metrics like FID, which measures quality and diversity based on the distance between the feature distributions of the original image and the resulting image, in contrast to "traditional distance", this refers to the distance between two multivariate probability distributions. In this case, the "point" is not just a single point in space, but the entire probability distribution. The Diffusion+Baseline model produces the lowest score indicating that the resulting image is of good quality where the important features of the original image can be replicated properly in the resulting image, and also the resulting image has good diversity or variation which is different from, where the FID with a low value indicates that the distribution of the resulting image features is similar to the distribution of the original image features. Meanwhile, based on visual evaluation, the Diffusion+Baseline model has the best new motif variations compared to other models, where combinations of patterns appear on motifs, even though these patterns are still taken from datasets.

In addition, the results can be analyzed that testing the model with Wassertein loss, although it shows an FID value nearly identical to that of StyleGAN2Loss, may not always yield high-quality images. This is because in Wassertein loss, the logits used to calculate the loss function are directly averaged to measure the distance between fake and original images, without the use of an activation function. This results in a wider range of variations in the produced samples, as even minor adjustments to the generator's weights and biases can lead to significant changes in the output. However, this could hinder the generator's ability to understand the dataset's intricacies, as these sudden changes could lead to an unstable learning phase.

In contrast, the StyleGAN2Loss mechanism applies a softplus activation function to the logits, which has the ability to refine its input. It's important to understand that even minor modifications in the input will correspondingly influence the output, and the same is true in reverse. This inherent characteristic contributes to a more stable training process, as slight adjustments to the generator's weights and biases won't result in a sudden decline in the quality of the generated samples. This allows the generator to achieve a more refined representation of the data \cite{karras2019style}.

The effectiveness of GAN is considerably increased by the incorporation of the diffusion approach. This is because the Diffusion approach uses a wide range of cutting-edge techniques, including Differential Augmentation \cite{shengyu2020}. Differential Augmentation offers a data-efficient method that increases the learning capacity of the model by modifying the data and samples generated before they are submitted to the discriminator. By adding changes to the dataset, differential augmentation can also assist avoid overfitting, diversify the training data, and improve training stability.

Additionally, there is a novel adaptive diffusion method that employs the same adaptive diffusion method to diffuse both observed and generated data. Every stage of the diffusion process uses a different noise-to-data ratio. As a result, the model is better equipped to manage noise changes, which improves its capacity to produce original and lifelike samples. Additionally, the discriminator is taught to distinguish between diffused created data and diffused real data at every stage of the diffusion process. Due to the environment this produces, the model's ability to adjust to different noise levels is much improved.

\section{Conclusion}\label{sec5:conclusion}
In the research that has been done, we have explored the use of the Diffusion-GAN method to synthesize batik motifs, we have conducted qualitative and quantitative tests on the results. Using the FID metric, we evaluate the results of the batik images produced by the model. The results obtained by the model show a value of around 29.045, indicating a good ability of the model to produce high-quality and diverse batik images

Besides that, we have also made a comparison between the Diffusion-GAN model and other approaches. In the experiments that have been carried out, we found that the Diffusion-GAN method is able to improve performance in producing batik motifs where high quality, diverse results are obtained and maintain the aesthetic appeal of the batik motifs themselves.

To achieve this result, we explored the use of loss in this study, such as Wassertein Loss and StyleGAN2Loss which use Non-Saturating loss, the results show Wassertein Loss produces patterns that tend to have new shapes, but the motifs tend to be not neatly arranged, compared to StyleGAN2Loss which produces neat motifs.

In addition, the dataset used in this research played a significant role in the success of the proposed method. The batik dataset we collected and utilized was diverse, encompassing various motif types and color variations. This diversity helped our model learn and emulate the distinctive characteristics of batik motifs more effectively. Moreover, the high quality of the dataset had a positive impact on the quality of the synthesized batik motifs.

Overall, this study contributes to the integration of Diffusion-GAN technology with traditional art and culture, particularly in the synthesis of batik motifs. The proposed method demonstrates its superiority in generating high-quality and authentic batik motifs. However, there is room for further development, such as improving the finesse and accuracy in generating finer batik motifs. With this research, we hope to inspire further advancements in the synthesis of batik motifs using machine learning-based approaches, ultimately supporting the sustainability and development of batik art and culture.

\section*{Declarations}


\begin{itemize}
 \item Funding \\
 This research received no specific grant from any funding agency in the public, commercial, or not-for-profit sectors.
 \item Availability of data and materials \\
 The datasets generated and/or analysed during the current study are available in {\href{https://github.com/octadion/diffusion-stylegan2-ada-pytorch}{https://github.com/octadion/diffusion-stylegan2-ada-pytorch}}
 \item Competing interests \\
 The authors declare that they have no competing interests
 \item Authors' contributions \\
 OO performed acquisition, analysis, interpretation of data, creation of new software used in the work, drafting the work, and substantively revised it. NY has made supervision, conception, design of the work, analysis, draft of the work, and substantively revised it;. DK supervised the works, has drafted the work, and substantively revised it. All authors read and approved the final manuscript
 \item Consent for publication \\
 Not applicable
 \item Acknowledgements \\
 Authors would thank Intelligent System Laboratory of Faculty of Computer Science, Brawijaya University and AI Center of Brawijaya University for providing high performance computing server.
\end{itemize}

\bibliography{sn-bibliography}


\begin{thebibliography}{36}
\ifx \bisbn   \undefined \def \bisbn  #1{ISBN #1}\fi
\ifx \binits  \undefined \def \binits#1{#1}\fi
\ifx \bauthor  \undefined \def \bauthor#1{#1}\fi
\ifx \batitle  \undefined \def \batitle#1{#1}\fi
\ifx \bjtitle  \undefined \def \bjtitle#1{#1}\fi
\ifx \bvolume  \undefined \def \bvolume#1{\textbf{#1}}\fi
\ifx \byear  \undefined \def \byear#1{#1}\fi
\ifx \bissue  \undefined \def \bissue#1{#1}\fi
\ifx \bfpage  \undefined \def \bfpage#1{#1}\fi
\ifx \blpage  \undefined \def \blpage #1{#1}\fi
\ifx \burl  \undefined \def \burl#1{\textsf{#1}}\fi
\ifx \doiurl  \undefined \def \doiurl#1{\url{https://doi.org/#1}}\fi
\ifx \betal  \undefined \def \betal{\textit{et al.}}\fi
\ifx \binstitute  \undefined \def \binstitute#1{#1}\fi
\ifx \binstitutionaled  \undefined \def \binstitutionaled#1{#1}\fi
\ifx \bctitle  \undefined \def \bctitle#1{#1}\fi
\ifx \beditor  \undefined \def \beditor#1{#1}\fi
\ifx \bpublisher  \undefined \def \bpublisher#1{#1}\fi
\ifx \bbtitle  \undefined \def \bbtitle#1{#1}\fi
\ifx \bedition  \undefined \def \bedition#1{#1}\fi
\ifx \bseriesno  \undefined \def \bseriesno#1{#1}\fi
\ifx \blocation  \undefined \def \blocation#1{#1}\fi
\ifx \bsertitle  \undefined \def \bsertitle#1{#1}\fi
\ifx \bsnm \undefined \def \bsnm#1{#1}\fi
\ifx \bsuffix \undefined \def \bsuffix#1{#1}\fi
\ifx \bparticle \undefined \def \bparticle#1{#1}\fi
\ifx \barticle \undefined \def \barticle#1{#1}\fi
\bibcommenthead
\ifx \bconfdate \undefined \def \bconfdate #1{#1}\fi
\ifx \botherref \undefined \def \botherref #1{#1}\fi
\ifx \url \undefined \def \url#1{\textsf{#1}}\fi
\ifx \bchapter \undefined \def \bchapter#1{#1}\fi
\ifx \bbook \undefined \def \bbook#1{#1}\fi
\ifx \bcomment \undefined \def \bcomment#1{#1}\fi
\ifx \oauthor \undefined \def \oauthor#1{#1}\fi
\ifx \citeauthoryear \undefined \def \citeauthoryear#1{#1}\fi
\ifx \endbibitem  \undefined \def \endbibitem {}\fi
\ifx \bconflocation  \undefined \def \bconflocation#1{#1}\fi
\ifx \arxivurl  \undefined \def \arxivurl#1{\textsf{#1}}\fi
\csname PreBibitemsHook\endcsname

\bibitem[\protect\citeauthoryear{Wronska-Friend}{2018}]{maria2018}
\begin{botherref}
\oauthor{\bsnm{Wronska-Friend}, \binits{M.}}:
Batik of java: Global inspiration.
Textile Society of America Symposium Proceedings 2018
(2018)
\end{botherref}
\endbibitem

\bibitem[\protect\citeauthoryear{Goodfellow
  et~al.}{2014}]{goodfellow2014generative}
\begin{botherref}
\oauthor{\bsnm{Goodfellow}, \binits{I.J.}},
\oauthor{\bsnm{Pouget-Abadie}, \binits{J.}},
\oauthor{\bsnm{Mirza}, \binits{M.}},
\oauthor{\bsnm{Xu}, \binits{B.}},
\oauthor{\bsnm{Warde-Farley}, \binits{D.}},
\oauthor{\bsnm{Ozair}, \binits{S.}},
\oauthor{\bsnm{Courville}, \binits{A.}},
\oauthor{\bsnm{Bengio}, \binits{Y.}}:
Generative adversarial networks.
arXiv preprint arXiv:1406.2661
(2014)
\end{botherref}
\endbibitem

\bibitem[\protect\citeauthoryear{Tero~Karras}{2019}]{karras2019style}
\begin{botherref}
\oauthor{\bsnm{Tero~Karras}, \binits{T.A.} \bsuffix{Samuli~Laine}}:
A style-based generator architecture for generative adversarial networks.
Computer Vision and Pattern Recognition.
(2019)
\end{botherref}
\endbibitem

\bibitem[\protect\citeauthoryear{Jonathan~Ho}{2020}]{ho2020denoising}
\begin{botherref}
\oauthor{\bsnm{Jonathan~Ho}, \binits{P.A.} \bsuffix{Ajay~Jain}}:
Denoising diffusion probabilistic models.
arXiv preprint arXiv:2006.11239v2
(2020)
\end{botherref}
\endbibitem

\bibitem[\protect\citeauthoryear{Martin~Arjovsky}{2017}]{arjovsky2017wgan}
\begin{botherref}
\oauthor{\bsnm{Martin~Arjovsky}, \binits{L.B.} \bsuffix{Soumith~Chintala}}:
Wassertein gan.
arXiv preprint arXiv:1701.07875v3
(2017)
\end{botherref}
\endbibitem

\bibitem[\protect\citeauthoryear{Gulrajani et~al.}{2017}]{gulrajani2017}
\begin{botherref}
\oauthor{\bsnm{Gulrajani}, \binits{I.}},
\oauthor{\bsnm{Ahmed}, \binits{F.}},
\oauthor{\bsnm{Arjovsky}, \binits{M.}},
\oauthor{\bsnm{Dumoulin}, \binits{V.}},
\oauthor{\bsnm{Courville}, \binits{A.}}:
Improved training of wasserstein gans.
arXiv:1704.00028v3 [cs.LG]
(2017)
\end{botherref}
\endbibitem

\bibitem[\protect\citeauthoryear{Fedus et~al.}{2018}]{fedus2018}
\begin{botherref}
\oauthor{\bsnm{Fedus}, \binits{W.}},
\oauthor{\bsnm{Rosca}, \binits{M.}},
\oauthor{\bsnm{Lakshminarayanan}, \binits{B.}},
\oauthor{\bsnm{Dai}, \binits{A.M.}},
\oauthor{\bsnm{Mohamed}, \binits{S.}},
\oauthor{\bsnm{Goodfellow}, \binits{I.}}:
Many paths to equilibrium: Gans do not need to decrease a divergence at every
  step.
arXiv:1710.08446v3 [stat.ML]
(2018)
\end{botherref}
\endbibitem

\bibitem[\protect\citeauthoryear{Karras et~al.}{2020}]{karras2020training}
\begin{botherref}
\oauthor{\bsnm{Karras}, \binits{T.}},
\oauthor{\bsnm{Laine}, \binits{S.}},
\oauthor{\bsnm{Aittala}, \binits{M.}},
\oauthor{\bsnm{Hellsten}, \binits{J.}},
\oauthor{\bsnm{Lehtinen}, \binits{J.}},
\oauthor{\bsnm{Aila}, \binits{T.}}:
Training generative adversarial networks with limited data.
NeurIPS
(2020)
\end{botherref}
\endbibitem

\bibitem[\protect\citeauthoryear{Meranggi et~al.}{2022}]{meranggi2022batik}
\begin{barticle}
\bauthor{\bsnm{Meranggi}, \binits{D.G.T.}},
\bauthor{\bsnm{Yudistira}, \binits{N.}},
\bauthor{\bsnm{Sari}, \binits{Y.A.}}:
\batitle{Batik classification using convolutional neural network with data
  improvements}.
\bjtitle{JOIV: International Journal on Informatics Visualization}
\bvolume{6}(\bissue{1}),
\bfpage{6}--\blpage{11}
(\byear{2022})
\end{barticle}
\endbibitem

\bibitem[\protect\citeauthoryear{Minarno et~al.}{2021}]{agus2021dcgan}
\begin{botherref}
\oauthor{\bsnm{Minarno}, \binits{A.E.}},
\oauthor{\bsnm{Mustaqim}, \binits{M.C.}},
\oauthor{\bsnm{Azhar}, \binits{Y.}},
\oauthor{\bsnm{Kusuma}, \binits{W.A.}},
\oauthor{\bsnm{Munarko}, \binits{Y.}}:
Deep convolutional generative adversarial network application in batik pattern
  generator.
International Conference on Information and Communication Technology (ICoICT)
(2021)
\end{botherref}
\endbibitem

\bibitem[\protect\citeauthoryear{Wang et~al.}{2022}]{zhendong2022}
\begin{botherref}
\oauthor{\bsnm{Wang}, \binits{Z.}},
\oauthor{\bsnm{Zheng}, \binits{H.}},
\oauthor{\bsnm{He}, \binits{P.}},
\oauthor{\bsnm{Chen}, \binits{W.}},
\oauthor{\bsnm{Zhou}, \binits{M.}}:
Diffusion-gan: Training gans with diffusion.
arXiv:2206.02262v3 [cs.LG]
(2022)
\end{botherref}
\endbibitem

\bibitem[\protect\citeauthoryear{Yohanes~Gultom}{2018}]{gultom2018batik}
\begin{botherref}
\oauthor{\bsnm{Yohanes~Gultom}, \binits{R.J.M.} \bsuffix{Aniati
  Murni~Arymurthy}}:
Batik classification using deep convolutional network transfer learning.
JURNAL ILMU KOMPURER DAN INFORMASI
(2018)
\end{botherref}
\endbibitem

\bibitem[\protect\citeauthoryear{Chrystian}{2023}]{chrys2023}
\begin{botherref}
\oauthor{\bsnm{Chrystian}, \binits{C.}}:
Itb-mbatik dataset
(2023)
\end{botherref}
\endbibitem

\bibitem[\protect\citeauthoryear{Zhao et~al.}{2020}]{shengyu2020}
\begin{botherref}
\oauthor{\bsnm{Zhao}, \binits{S.}},
\oauthor{\bsnm{Liu}, \binits{Z.}},
\oauthor{\bsnm{Lin}, \binits{J.}},
\oauthor{\bsnm{Zhu}, \binits{J.-Y.}},
\oauthor{\bsnm{Han}, \binits{S.}}:
Differentiable augmentation for data-efficient gan training.
Advances in Neural Information Processing Systems, 33: 7559–7570
(2020)
\end{botherref}
\endbibitem

\bibitem[\protect\citeauthoryear{Karras et~al.}{2020}]{karras2020analyze}
\begin{botherref}
\oauthor{\bsnm{Karras}, \binits{T.}},
\oauthor{\bsnm{Laine}, \binits{S.}},
\oauthor{\bsnm{Aittala}, \binits{M.}},
\oauthor{\bsnm{Hellsten}, \binits{J.}},
\oauthor{\bsnm{Lehtinen}, \binits{J.}},
\oauthor{\bsnm{Aila}, \binits{T.}}:
Analyzing and improving the image quality of stylegan.
CVPR
(2020)
\end{botherref}
\endbibitem

\bibitem[\protect\citeauthoryear{Chrystian}{2023}]{christian2023batikgan}
\begin{botherref}
\oauthor{\bsnm{Chrystian}, \binits{W.}}:
Sp-batikgan: An efficient generative adversarial network for symmetric pattern
  generation.
cs.CV
(2023)
\end{botherref}
\endbibitem

\bibitem[\protect\citeauthoryear{Karras et~al.}{2019}]{karras2019analyze2}
\begin{botherref}
\oauthor{\bsnm{Karras}, \binits{T.}},
\oauthor{\bsnm{Laine}, \binits{S.}},
\oauthor{\bsnm{Aittala}, \binits{M.}},
\oauthor{\bsnm{Hellsten}, \binits{J.}},
\oauthor{\bsnm{Lehtinen}, \binits{J.}},
\oauthor{\bsnm{Aila}, \binits{T.}}:
Analyzing and improving the image quality of stylegan.
CoRR, vol. abs/1912.0
(2019)
\end{botherref}
\endbibitem

\bibitem[\protect\citeauthoryear{anniemi et~al.}{2019}]{tuomas2019improved}
\begin{botherref}
\oauthor{\bsnm{anniemi}, \binits{T.K.}},
\oauthor{\bsnm{Karras}, \binits{T.}},
\oauthor{\bsnm{Laine}, \binits{S.}},
\oauthor{\bsnm{Lehtinen}, \binits{J.}},
\oauthor{\bsnm{Aila}, \binits{T.}}:
Improved precision and recall metric for assessing generative models.
CoRR, vol. abs/1904.06991
(2019)
\end{botherref}
\endbibitem

\bibitem[\protect\citeauthoryear{Zhou et~al.}{2018}]{yzhou2018non}
\begin{botherref}
\oauthor{\bsnm{Zhou}, \binits{Y.}},
\oauthor{\bsnm{Zhu}, \binits{Z.}},
\oauthor{\bsnm{Bai}, \binits{X.}},
\oauthor{\bsnm{Lischinski}, \binits{D.}},
\oauthor{\bsnm{Cohen-Or}, \binits{D.}},
\oauthor{\bsnm{Huang}, \binits{H.}}:
Non-stationary texture synthesis by adversarial expansion.
cs.GR
(2018)
\end{botherref}
\endbibitem

\bibitem[\protect\citeauthoryear{Xian~Wu}{2017}]{xianwu2017}
\begin{botherref}
\oauthor{\bsnm{Xian~Wu}, \binits{P.H.} \bsuffix{Kun~Xu}}:
A survey of image synthesis and editing with generative adversarial networks.
Tsinghua Science and Technology
(2017)
\end{botherref}
\endbibitem

\bibitem[\protect\citeauthoryear{Naeem et~al.}{2020}]{muhammad2020}
\begin{botherref}
\oauthor{\bsnm{Naeem}, \binits{M.F.}},
\oauthor{\bsnm{Oh}, \binits{S.J.}},
\oauthor{\bsnm{Uh}, \binits{Y.}},
\oauthor{\bsnm{Choi}, \binits{Y.}},
\oauthor{\bsnm{Yoo}, \binits{J.}}:
Reliable fidelity and diversity metrics for generative models
(2020)
\end{botherref}
\endbibitem

\bibitem[\protect\citeauthoryear{Yu et~al.}{2021}]{ningyu2021}
\begin{botherref}
\oauthor{\bsnm{Yu}, \binits{N.}},
\oauthor{\bsnm{Liu}, \binits{G.}},
\oauthor{\bsnm{Dundar}, \binits{A.}},
\oauthor{\bsnm{Tao}, \binits{A.}},
\oauthor{\bsnm{Catanzaro}, \binits{B.}},
\oauthor{\bsnm{Davis}, \binits{L.}},
\oauthor{\bsnm{Fritz}, \binits{M.}}:
Dual contrastive loss and attention for gans.
IEEE International Conference on Computer Vision (ICCV)
(2021)
\end{botherref}
\endbibitem

\bibitem[\protect\citeauthoryear{Bellemare et~al.}{2017}]{marc2017}
\begin{botherref}
\oauthor{\bsnm{Bellemare}, \binits{M.G.}},
\oauthor{\bsnm{Danihelka}, \binits{I.}},
\oauthor{\bsnm{Dabney}, \binits{W.}},
\oauthor{\bsnm{Mohamed}, \binits{S.}},
\oauthor{\bsnm{Lakshminarayanan}, \binits{B.}},
\oauthor{\bsnm{Hoyer}, \binits{S.}},
\oauthor{},
\oauthor{\bsnm{Munos}, \binits{R.}}:
The cramer distance as a solution to biased wasserstein gradients.
arXiv preprint arXiv:1705.10743
(2017)
\end{botherref}
\endbibitem

\bibitem[\protect\citeauthoryear{Brock et~al.}{2018}]{andrew2018}
\begin{botherref}
\oauthor{\bsnm{Brock}, \binits{A.}},
\oauthor{\bsnm{Donahue}, \binits{J.}},
\oauthor{\bsnm{Simonyan}, \binits{K.}}:
Large scale gan training for high fidelity natural image synthesis.
arXiv preprint arXiv:1809.11096
(2018)
\end{botherref}
\endbibitem

\bibitem[\protect\citeauthoryear{Deshpande et~al.}{2018}]{ishan2018}
\begin{botherref}
\oauthor{\bsnm{Deshpande}, \binits{I.}},
\oauthor{\bsnm{Zhang}, \binits{Z.}},
\oauthor{\bsnm{Schwing}, \binits{A.G.}}:
Generative modeling using the sliced wasserstein distance.
Proceedings of the IEEE conference on computer vision and pattern recognition,
  pages 3483–3491
(2018)
\end{botherref}
\endbibitem

\bibitem[\protect\citeauthoryear{Dhariwal and Nichol}{2021}]{prafulla2021}
\begin{botherref}
\oauthor{\bsnm{Dhariwal}, \binits{P.}},
\oauthor{\bsnm{Nichol}, \binits{A.Q.}}:
Diffusion models beat gans on image synthesis.
A. Beygelzimer, Y. Dauphin, P. Liang, and J. Wortman Vaughan, editors, Advances
  in Neural Information Processing Systems, 2021. URL https://openreview.net/
  forum?id=AAWuCvzaVt
(2021)
\end{botherref}
\endbibitem

\bibitem[\protect\citeauthoryear{Heusel et~al.}{2017}]{heusel2017}
\begin{botherref}
\oauthor{\bsnm{Heusel}, \binits{M.}},
\oauthor{\bsnm{Ramsauer}, \binits{H.}},
\oauthor{\bsnm{Unterthiner}, \binits{T.}},
\oauthor{\bsnm{Nessler}, \binits{B.}},
\oauthor{\bsnm{Hochreiter}, \binits{S.}}:
Gans trained by a two time-scale update rule converge to a local nash
  equilibrium.
Advances in Neural Information Processing Systems, pages 6626–6637
(2017)
\end{botherref}
\endbibitem

\bibitem[\protect\citeauthoryear{Kong and Ping}{2021}]{zhifeng2021}
\begin{botherref}
\oauthor{\bsnm{Kong}, \binits{Z.}},
\oauthor{\bsnm{Ping}, \binits{W.}}:
On fast sampling of diffusion probabilistic models.
arXiv preprint arXiv:2106.00132
(2021)
\end{botherref}
\endbibitem

\bibitem[\protect\citeauthoryear{Liu et~al.}{2020}]{bingchen2020}
\begin{botherref}
\oauthor{\bsnm{Liu}, \binits{B.}},
\oauthor{\bsnm{Zhu}, \binits{Y.}},
\oauthor{\bsnm{Song}, \binits{K.}},
\oauthor{\bsnm{Elgammal}, \binits{A.}}:
Towards faster and stabilized gan training for high-fidelity few-shot image
  synthesis.
International Conference on Learning Representations
(2020)
\end{botherref}
\endbibitem

\bibitem[\protect\citeauthoryear{Mescheder et~al.}{2017}]{lars2017}
\begin{botherref}
\oauthor{\bsnm{Mescheder}, \binits{L.}},
\oauthor{\bsnm{Nowozin}, \binits{S.}},
\oauthor{\bsnm{Geiger}, \binits{A.}}:
The numerics of gans.
Advances in neural information processing systems, 30
(2017)
\end{botherref}
\endbibitem

\bibitem[\protect\citeauthoryear{Mescheder et~al.}{2018}]{lars2018}
\begin{botherref}
\oauthor{\bsnm{Mescheder}, \binits{L.}},
\oauthor{\bsnm{Geiger}, \binits{A.}},
\oauthor{\bsnm{Nowozin}, \binits{S.}}:
Which training methods for gans do actually converge?
International conference on machine learning, pages 3481–3490. PMLR
(2018)
\end{botherref}
\endbibitem

\bibitem[\protect\citeauthoryear{Roth et~al.}{2017}]{kevin2017}
\begin{botherref}
\oauthor{\bsnm{Roth}, \binits{K.}},
\oauthor{\bsnm{Lucchi}, \binits{A.}},
\oauthor{\bsnm{Nowozin}, \binits{S.}},
\oauthor{\bsnm{Hofmann}, \binits{T.}}:
Stabilizing training of generative adversarial networks through regularization.
Advances in neural information processing systems, 30
(2017)
\end{botherref}
\endbibitem

\bibitem[\protect\citeauthoryear{Xiao et~al.}{2022}]{zhiseng2022}
\begin{botherref}
\oauthor{\bsnm{Xiao}, \binits{Z.}},
\oauthor{\bsnm{Kreis}, \binits{K.}},
\oauthor{\bsnm{Vahdat}, \binits{A.}}:
Tackling the generative learning trilemma with denoising diffusion gans.
International Conference on Learning Representations, 2022. URL
  https://openreview.net/forum?id=JprM0p-q0Co
(2022)
\end{botherref}
\endbibitem

\bibitem[\protect\citeauthoryear{Tran et~al.}{2021}]{trungtran2021}
\begin{botherref}
\oauthor{\bsnm{Tran}, \binits{N.-T.}},
\oauthor{\bsnm{Tran}, \binits{V.-H.}},
\oauthor{\bsnm{Nguyen}, \binits{N.-B.}},
\oauthor{\bsnm{Nguyen}, \binits{T.-K.}},
\oauthor{\bsnm{Cheung}, \binits{N.-M.}}:
On data augmentation for gan training.
IEEE Transactions on Image Processing, 30:1882–1897
(2021)
\end{botherref}
\endbibitem

\bibitem[\protect\citeauthoryear{San-Roman et~al.}{2021}]{robin2021}
\begin{botherref}
\oauthor{\bsnm{San-Roman}, \binits{R.}},
\oauthor{\bsnm{Nachmani}, \binits{E.}},
\oauthor{\bsnm{Wolf}, \binits{L.}}:
Noise estimation for generative diffusion models.
arXiv preprint arXiv:2104.02600
(2021)
\end{botherref}
\endbibitem

\bibitem[\protect\citeauthoryear{Zheng et~al.}{2022}]{huangjie2022}
\begin{botherref}
\oauthor{\bsnm{Zheng}, \binits{H.}},
\oauthor{\bsnm{He}, \binits{P.}},
\oauthor{\bsnm{Chen}, \binits{W.}},
\oauthor{\bsnm{Zhou}, \binits{M.}}:
Truncated diffusion probabilistic models.
arXiv preprint arXiv:2202.09671
(2022)
\end{botherref}
\endbibitem

\end{thebibliography}
\nocite{karras2020analyze}
\nocite{karras2020training} 
\nocite{christian2023batikgan}
\nocite{karras2019analyze2}
\nocite{tuomas2019improved}
\nocite{yzhou2018non}
\nocite{xianwu2017}
\nocite{muhammad2020}
\nocite{ningyu2021}
\nocite{marc2017}
\nocite{andrew2018}
\nocite{ishan2018}
\nocite{prafulla2021}
\nocite{heusel2017}
\nocite{zhifeng2021}
\nocite{bingchen2020}
\nocite{lars2017}
\nocite{lars2018}
\nocite{kevin2017}
\nocite{zhiseng2022}
\nocite{trungtran2021}
\nocite{robin2021}
\nocite{huangjie2022}
\end{document}